\pdfoutput=1

\documentclass[11pt]{article}

\usepackage{emnlp2022}

\usepackage{times}
\usepackage{latexsym}
\usepackage{xspace}
\usepackage{tabularray}
\usepackage{amsmath}
\newcommand{\base}{\ensuremath{\mathcal{B}}\xspace}
\newcommand{\targ}{\ensuremath{\mathcal{T}}\xspace}

\usepackage{subcaption}
\newcommand{\M}{\ensuremath{\mathcal{M}}\xspace}

\newcommand{\baseel}[1]{\ensuremath{b_{#1}}\xspace}
\newcommand{\targel}[1]{\ensuremath{t_{#1}}\xspace}
\newcommand{\bi}{\ensuremath{b_{i}}\xspace}

\newcommand{\bj}{\ensuremath{b_{j}}\xspace}

\newcommand{\tl}{\ensuremath{t_{l}}\xspace}

\newcommand{\cossimstar}{\ensuremath{\textit{sim}^*}\xspace}
\newcommand{\simrel}{{\textit{sim}}\xspace}
\newcommand{\rel}{\ensuremath{\mathcal{R}}\xspace}

\newcommand{\dnote}[1]{\textcolor{blue}{$\ll$\textsf{#1 | Dafna}$\gg$}}
\newcommand{\onote}[1]{\textcolor{red}{$\ll$\textsf{#1 | Oren}$\gg$}}

\DeclareMathOperator*{\argmax}{arg\,max}

\newcommand{\remove}[1]{}

\usepackage{times}
\usepackage{latexsym}

\usepackage[T1]{fontenc}

\usepackage[utf8]{inputenc}

\usepackage{microtype}

\usepackage{inconsolata}
\usepackage{paralist}

\newcommand{\xhdr}[1]{\vspace{1mm}\noindent{{\bf #1.}}} 

\usepackage{graphicx}

\title{Life is a Circus and We are the Clowns: \\ Automatically Finding Analogies between Situations and Processes}

\author{Oren Sultan \\
 The Hebrew University of Jerusalem \\
  \texttt{oren.sultan@mail.huji.ac.il} \\\And
  Dafna Shahaf \\
  The Hebrew University of Jerusalem \\
  \texttt{dshahaf@cs.huji.ac.il} \\
}

\begin{document}
\maketitle
\begin{abstract}


Analogy-making gives rise to reasoning, abstraction, flexible categorization and counterfactual 
inference -- abilities lacking in even the best AI systems today. Much research has suggested that analogies are key to non-brittle systems that can adapt to new domains. Despite their importance, analogies received little attention in the NLP community, with most research focusing on simple word analogies. Work that tackled more complex analogies relied heavily on manually constructed, hard-to-scale input representations. In this work, we explore a more realistic, challenging setup: our input is a pair of natural language \emph{procedural texts}, describing a situation or a process (e.g., how the heart works/how a pump works). Our goal is to automatically extract entities and their relations from the text and find a mapping between the different domains based on \emph{relational similarity} (e.g., blood is mapped to water). 
We develop an interpretable, scalable algorithm and demonstrate that it identifies the correct mappings 87\% of the time for \emph{procedural texts} and 94\% for \emph{stories} from cognitive-psychology literature. We show it can extract analogies from a large dataset of procedural texts, achieving 79\% precision (analogy prevalence in data: 3\%). Lastly, we demonstrate that our algorithm is robust to paraphrasing the input texts\footnote{{Data and code are available in our GitHub repository \url{ https://github.com/orensul/analogies_mining}}}.
\end{abstract}

\section{Introduction}


\begin{figure*}[t]
\includegraphics[width=.99\textwidth]{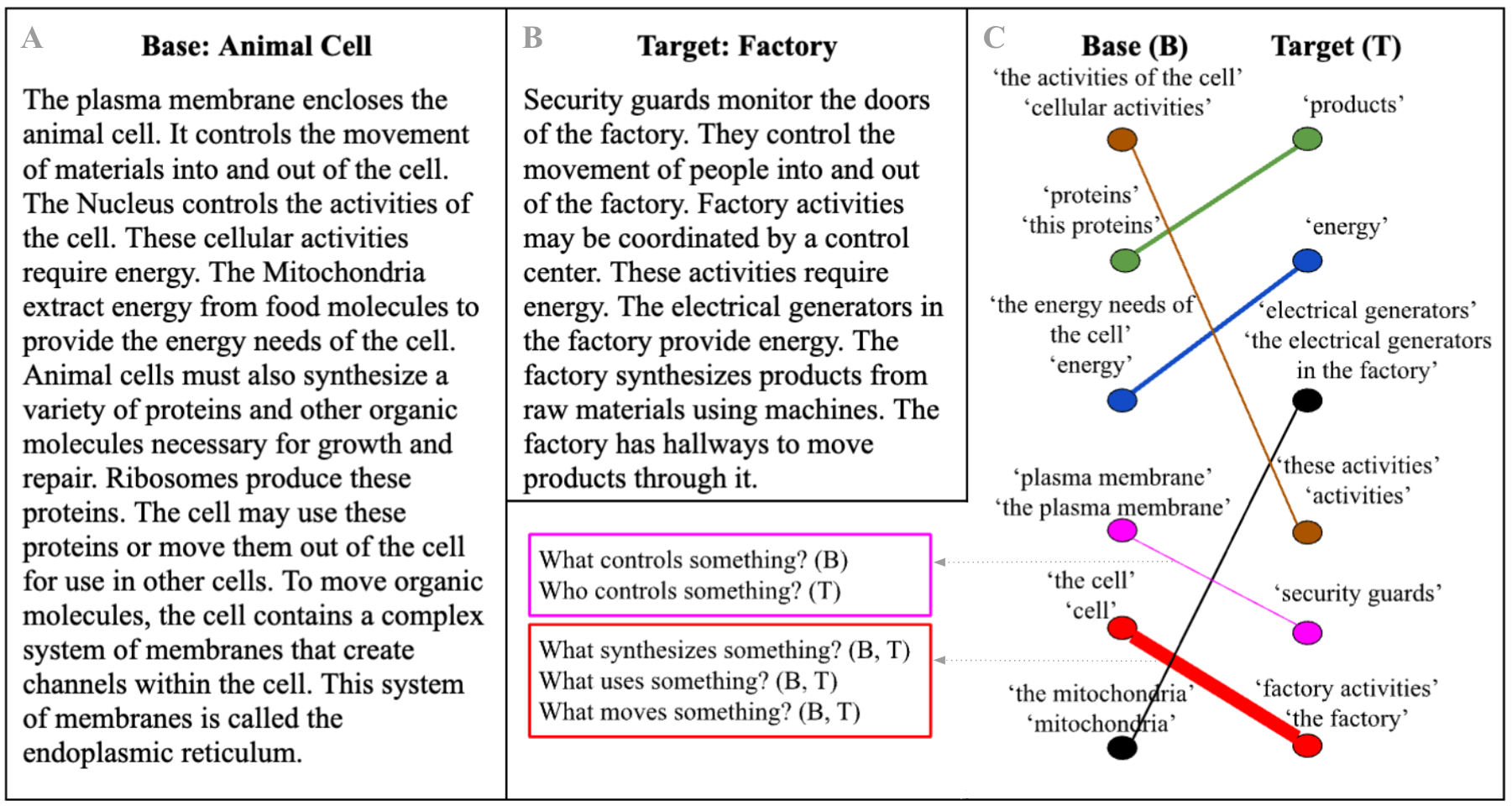}
\caption{A+B: Example input -- two analogous texts, describing the animal cell (base) and factory (target), adapted from \citet{mikelskis2008four}. 
C: Our algorithm’s output. The nodes are entities (clusters of text spans; for the sake of presentation, we show up to two spans per cluster). Edge width represents similarity between entities in terms of the roles they play in the text. For example, the boxes on the left illustrate the similar roles associated with the red and the pink entities.  
All the mappings our algorithm found are correct, but two are missing (ribosomes/machines and endoplasmic reticulum/hallways). Showing the mapping along with its justification (the similar roles) renders our output \textit{easy to interpret}.
}
\label{fig:animal_cell_factory_best_solution}
\end{figure*}

The ability to find parallels across diverse domains and transfer ideas across them is one of the pinnacles of human cognition. The \emph{analogous reasoning} process allows us to abstract information, form flexible concepts and solve problems based on our previous experience \citep{minsky1988society,hofstadter2013surfaces,holyoak1984analogical} -- abilities that current AI systems still lack. 
Many researchers have suggested that analogy-making is essential for engineering non-brittle AI that can robustly generalize and adapt to diverse domains \citep{Mitchell2021AbstractionAA}.


Surprisingly, despite analogy's important role in the way humans understand language, the problem of recognizing analogies has received relatively little attention in NLP. Most works have focused on SAT-type of analogies (``\emph{a} to \emph{b} is like \emph{c} to \emph{d}''), with
%
%
recent works \citep{linzen-2016-issues, ushio2021bert, NEURIPS2020_1457c0d6}
showing that models 
still struggle with abstract, loosely defined relations. 

In this work, we focus on a different type of analogies: analogies between \emph{situations} or \emph{processes}. Here the input is two domains  (e.g., heart and pump) and the goal is to map objects from the base domain to objects from the target domain. Importantly, the  mapping should rely on a common \emph{relational structure} rather than object attributes, making it challenging for NLP methods.

The most influential work in this line of research is  Structure Mapping Theory (SMT)  \citep{GENTNER1983155}. In SMT and much of its follow-up work \citep{falkenhainer1989structure,turney2008latent, forbus2011cogsketch}, domain descriptions are sets of statements in a highly structured language, e.g.,

\begin{footnotesize}
\begin{verbatim}
CAUSE(PULL(piston),CAUSE(GREATER(PRESSURE(water),
PRESSURE(pipe)),FLOW(water,pipe)))
\end{verbatim}
\end{footnotesize}
%
%
%
Many have argued that too much human creativity is required to construct these inputs, and the analogy is already expressed in them \cite{chalmers1992high}. They are also brittle and hard-to-scale. 



In our work, we explore a more realistic, challenging setup. Our input is a pair of two \emph{procedural texts}, describing a situation or a process, expressed in natural language. 
We develop an algorithm to automatically extract entities and their relations from the text and find \emph{a mapping} between the different domains based on \emph{relational similarity}.

For example, the two texts in Figure~\ref{fig:animal_cell_factory_best_solution} 
explain the animal cell to students through an analogy to a factory (adapted from  \citet{mikelskis2008four})\footnote{Note that this example is from a textbook, given for simplicity of presentation, and thus the texts are particularly well-aligned. Our method can handle much more variation.}. Note that in our setting entities and relations are not given upfront, but need to be extracted; they can appear multiple times, often expressed in different terms. Also, entities can play multiple roles throughout the text.

Figure~\ref{fig:animal_cell_factory_best_solution}{C} shows our algorithm's output. Nodes are entities (clusters of text spans -- mostly correct, except for ``factory activities''), and edge width represents similarity in terms of roles entities play.
For example, the boxes on the left illustrate the similar roles associated with the red and the pink entities (represented as questions), showing that ``plasma membrane'' and ``security guards'' share one role (control something), and ``cell'' and ``factory'' share three. Importantly, including the justification as part of the output renders our algorithm \textit{interpretable}.
We note that all the mappings our algorithm found are correct, but two are missing (ribosomes/machines and endoplasmic reticulum/hallways). 
\textbf{Our contributions are:}
\begin{compactitem}
    \item We present a novel setting in computational analogy -- mapping between \emph{procedural texts} expressed in natural language.
    We develop a scalable, interpretable method to find mappings based on \emph{relational similarity}. 
    \item Our method identifies the correct mappings 87\% of the time for \emph{procedural texts} from ProPara dataset \citep{proparNaacl2018}. Although not designed for them, our method achieves 94\% precision on cognitive-psychology \emph{stories} \citep{gentner1993roles,ichienVerbalAnalogyProblem2020}.
    \item We demonstrate our method can be used to mine analogies in the ProPara dataset, achieving 79\% precision at the top of the ranking, when analogy prevalence in data is $\sim 3\%$.
    \item We show our method is robust to \emph{paraphrasing} the input texts. 
    \item We release data and code, including the found mappings {at \url{ https://github.com/orensul/analogies_mining}}. 
\end{compactitem}
We hope this work will pave the way for further NLP work on computational analogy. 

\section{Problem Formulation}

Our framework is based on Gentner's structure mapping theory (SMT) \citep{GENTNER1983155}. The central idea of SMT is that an analogy is a mapping of objects from a base domain \base into a target domain \targ that is based on a common \emph{relational structure}, rather than object attributes. 
In our setup, the input is two \emph{procedural texts} describing a situation or a process, expressed in natural language. 

\xhdr{Entities}
First, we need to extract entities from the texts. Let \base = $\{b_{1},...,b_{n}\}$
and \targ = $\{t_{1},...,t_{m}\}$ be the entities in the base and the target, respectively. In our setting, entities are noun phrases. For example, in the animal cell (Figure~\ref{fig:animal_cell_factory_best_solution}), some entities include plasma membrane, animal cell, nucleus, energy, mitochondria, proteins, ribosomes. 

\xhdr{Relations}
Let \rel be a set of relations. A relation is a set of ordered entity pairs. 
In this work, we focus on verbs from the input text, but other formulations are possible. 
Intuitively,
relations should capture that {the mitochondria} \emph{provides} {energy} to {the cell} (in \base), just like {electrical generators} \emph{provide} energy to {the factory} (in \targ). 
Slightly abusing notation, let $\rel(e_1,e_2) \subseteq 2^\rel$ be the set of relations between  $e_1$ and $e_2$. 
For example, $\rel(cell, proteins)$ contains \{synthesize, use, move\}. 
Note that \rel is an asymmetric function, as the order matters.


\xhdr{Similarity}
Let \simrel be a similarity metric between two sets of relations, $\simrel : 2^\rel\times2^\rel\to[0, \infty)$. 
Intuitively, we want similarity to be high if the two sets \emph{share many distinct} relations. 
For example, \{provide, destroy\}, should be more similar to \{supply, ruin\} than to \{destroy, ruin\} as the last set does not include any relation similar to provide.
Given a pair of entities $b_i,b_j\in\base$  and a pair of entities $t_k,t_l\in\targ$, we define a similarity function measuring how similar these pairs are, in terms of the relations between them. Since \simrel is asymmetric, we consider both possible orderings:
%
\begin{align}
    \cossimstar(b_i,b_j,  t_k, & t_l) = 
    \simrel(\rel(b_i,b_j), \rel(t_k,t_l))  \nonumber \\ 
     + & \simrel(\rel(b_j,b_i), \rel(t_l,t_k))
\label{eq:pairs_rel_sim}
\end{align}
%
%
%
%
\xhdr{Objective}
Our goal is to find a mapping function $\ensuremath{\mathcal{M}}: \base \to \targ \cup \bot$ that maps entities from base to target. Mapping into $\bot$ means the entity was not mapped. The mapping should be \emph{consistent} -- no two base entities can be mapped to the same entity. 
We look for a mapping that maximizes the relational similarity between mapped pairs:

\begin{equation}
    \mathcal{M}^*=\argmax_{\mathcal{M}}\!\!\sum_{\substack{j\in [1,n-1]\\ i\in[j+1,n]}}\!\!\!\cossimstar(b_j,b_i,\mathcal{M}(b_j),\mathcal{M}(b_i))
\nonumber
\end{equation}

\noindent If $b_i$ or $b_j$ maps to $\bot$, $\cossimstar$ is defined to be 0.

\label{sec:problemDefinition}

\section{Analogous Matching Algorithm}
Our goal in this section is to find the best mapping between \base and \targ. Our algorithm consists of four phases: we begin with a basic \textbf{text processing} {(Section \ref{subsec:textProcessing})}. Then, we \textbf{extract potential entities and relations}{
(Section \ref{subsec:structureExtraction})}. Since entities can be referred to in multiple ways, we next \textbf{cluster} {the entities (Section \ref{subsec:agglomerativeClustering})}. 
Finally, we find a {\bf mapping} between clusters from \base and \targ {(Section \ref{subsec:findMappings})}.

{We note that our goal in this paper is to present a new task and find a reasonable model for it; many other architectures and design choices are possible and could be explored in future work.}

\subsection{Text Processing}
\label{subsec:textProcessing}
We begin by chunking the sentences in the input. As our next step is structure extraction, we first want to resolve pronouns. We apply a lightweight \emph{co-reference} model \citep{Kirstain2021CoreferenceRW} which generates clusters of strings (e.g, \emph{the plasma membrane}, \emph{plasma membrane}, \emph{it}) and replace all pronouns by a representative from their cluster -- the shortest string which is not a pronoun or a verb.



\subsection{Structure Extraction}
\label{subsec:structureExtraction}



Analogy is based on relational similarity; thus, we now extract relations from the text. This naturally falls under Semantic Role Labeling (SRL) \citep{10.1162/089120102760275983} -- identifying the underlying relationship that words have with the main verb in a clause. 
In particular, we chose to use QA-SRL \citep{fitzgerald2018large}. This model receives a sentence as input and outputs questions 
and answers about the sentence. 
See Table~\ref{tab:QA} for example questions and answers. {Intuitively, the spans identified by QA-SRL as answers form the entities, and similar questions that appear in both domains (e.g., ``what provides something?'') indicate that the two entities (mitochondria, generators) may play similar roles.}

We chose to use QA-SRL since it allows the questions themselves to define the set of roles, with no predefined frame or thematic role ontologies. 
Recent studies show that QA-SRL achieves 90\% coverage of PropBank arguments, while capturing much implicit information that is often missed by traditional SRL schemes \citep{roit2020controlled}. 

We focus on questions likely to capture useful relations for our task. We filter out ``When'' and ``Why'' questions, ``Be'' verbs, and questions and answers with a low probability (see Appendix~\ref{sec:ignore_qa}). 

\begin{table*}[t!]
\begin{tblr}{hlines}
\textbf{Text} & \textbf{Verb} & \textbf{Question} & \textbf{Answer} \\
The animal cell  & provide & {what provides something?\\what provides something?\\ what provides something?\\what does something provide?} & {the mitochondria\\mitochondria\\food molecules\\the energy needs of the cell} \\
\hline
The factory & provide & {what provides something?\\what provides something?\\what does something provide?} & {the electrical generators\\the electrical generators in the factory\\energy} \\
\end{tblr}
\caption{Snippet from QA-SRL output on our cell/factory texts. 
}
\label{tab:QA}
\end{table*}







\subsection{Clustering Entities}
\label{subsec:agglomerativeClustering}

In classical computational analogy work, entities are explicitly given, each with a unique name (``cell''). However, in our input, entities are often referred to in different ways (\emph{``the animal cell''}, \emph{``the cell''}, \emph{``cell''}), which might confuse the mapping algorithm. Therefore, in this step we 
merge those different phrasings, resulting in a new, more refined set of entities. 
Since we do not know in advance the number of clusters, we use Agglomerative Clustering \citep{Zepeda-Mendoza2013}. {We manually fine-tuned the linkage threshold that determines the number of clusters} (see Appendix~\ref{subsec:other_params} for details). 
We denote the resulting clusters of entities as \base = $\{b_{1},...,b_{n}\}$ 
and \targ = $\{t_{1},...,t_{m}\}$. 
Figure~\ref{fig:animal_cell_clustering} shows the output clusters for the animal cell text. Most entities are identified correctly, although not all (e.g., clusters 5 and 7 should merge).

\subsection{Find Mappings}
\label{subsec:findMappings}


Our problem definition (Equation \ref{eq:pairs_rel_sim}) assumes we know all relations between the entities. However, our extraction algorithm is not perfect -- e.g., it cannot detect relations described across sentences, or using complex references. Consider these sentences (slight paraphrases of the original texts): 


\begin{center}
    \it 
``Animal cells must also produce proteins and other organic molecules necessary for growth and repair. Ribosomes are used for this process'' /
``The factory synthesizes products from raw materials using machines'' 
\end{center}


Ideally, we would like to infer that \emph{ribosomes} produce \emph{proteins} and \emph{machines} synthesize \emph{products}. QA-SRL only gives us partial information, but it is still useful. For example, both \emph{proteins} and \emph{products} are associated with similar questions (what is produced?, what is synthesized?), hinting that they might play similar roles.
Thus, we propose a \emph{heuristic approach} to approximate 
Equation \ref{eq:pairs_rel_sim}. 

Intuitively, the similarity score between two entities $\bi,\targel{k}$ is high if the similarity between their associated \emph{questions} is high (for example,  
\emph{cell}  and \emph{factory}  have multiple {distinct similar questions}). 
We define this as the sum of cosine distances over their associated questions' SBERT \citep{reimers-2019-sentence-bert} embeddings. We filter out distances below a similarity threshold (see Appendix~\ref{subsec:cosine_threshold}).
If there exists a sentence involving \bi and \baseel{j} and another sentence involving \targel{k} and \targel{l}, such that those sentences were used \emph{both} for mapping \bi to \targel{k} and \baseel{j} to \targel{l} using \emph{the same verb}, a complete \emph{relation} was found. In this case, we increase the score of both mappings. There are multiple ways to do so; we found that a simple schema of adding a constant $\alpha$ to the similarity of $(\bi,\targel{k})$ and $(\bj, \targel{l})$ works well in practice (see Appendix~\ref{subsec:other_params} for parameters).

We observe that questions are mostly of similar length (in ProPara, $\sim$1/3 of the questions have 3 words, $\sim$1/3 have 4 words, $\sim$1/6 have 2 words and $\sim$1/6 have 5 words). Note that the entities are not part of the questions.




\begin{figure}[t!]
\fbox{\parbox[c]{\linewidth}{
\textbf{1)} 'system of membranes', 'a complex system of membranes', 'this system of membranes', 'membranes', 'complex system of membranes'. 
\textbf{2)} 'food molecules', 'organic molecules'. 
\textbf{3)} 'energy', 'the energy needs of the cell'.
\textbf{4)} 'these proteins', 'proteins'.
\textbf{5)} 'animal cells', 'animal cell', 'the animal cell'.
\textbf{6)} 'these cellular activities', 'cellular activities', 
'the activities of the cell'.
\textbf{7)} 'the cell', 'cell'.
\textbf{8)} 'nucleus', 'the nucleus'.
\textbf{9)} 'endoplasmic reticulum', 'the endoplasmic reticulum'.
\textbf{10)} 'mitochondria', 'the mitochondria'.
\textbf{11)} 'channels'.
\textbf{12)} 'the plasma membrane', 'plasma membrane'.
\textbf{13)} 'the activities'.
\textbf{14)} 'ribosomes'.
\textbf{15)} 'movement of materials', 'the movement of materials'.
}}
\caption{The result of agglomerative clustering on animal cell text. Most of the entities were identified correctly, but not all (e.g., clusters 5 and 7 should merge).}
\label{fig:animal_cell_clustering}
\end{figure}

\xhdr{Beam Search}
\label{subsec:beam_search}
After computing all similarities, we use beam search to find the mapping $\ensuremath{\mathcal{M}}^{*}$
(see Appendix~\ref{subsec:other_params} for parameters).

Figure~\ref{fig:animal_cell_factory_best_solution} shows our algorithm's top mapping for the factory/cell example. Edge weights represent similarity. 
All the mappings our algorithm found are correct, but two are missing (ribosomes/machines and endoplasmic reticulum/hallways).
Being able to show the user the {relations} that led to the output mapping renders our method easy to interpret.
We name it {\bf Find Mappings by Questions (FMQ)}. 


\section{Experiments}

Our research questions are as follows:
\begin{compactitem}
\item \textbf{RQ1}: Can we leverage our algorithm for \emph{retrieving} analogies from a large dataset of procedural texts?
\item \textbf{RQ2}: Does our algorithm produce the correct mapping solution?
\item \textbf{RQ3}: Is our algorithm \emph{robust} to paraphrasing the input texts?
\end{compactitem}




\label{subsec:researchQuestions}

\hfill\break
We chose to test our ideas on 
the {\bf ProPara} dataset \citep{proparNaacl2018} of crowdsourced paragraphs describing processes. Prompts (e.g., \textquotedblleft What happens during photosynthesis?\textquotedblright) were given to 1-6 workers each. 
We used the  ProPara training set (390 paragraphs). 
See Appendix~\ref{subsec:propara_paragraphs} for two example paragraphs and our algorithm's mapping. Experiments run from a laptop. Runtime is few seconds for a pair of paragraphs on average.

\subsection{Mining Analogies}
\label{subsec:analogiesMining}

\begin{table*}[t]
\centering
\begin{tabular}{lll}
\hline
\textbf{Prompt in Base} & \textbf{Prompt in Target} & \textbf{Analogy Type}\\
\hline
Describe how oxygen reaches cells & What do lungs do? & \emph{Self analogy}\\

How does rain form? & How does snow form? & \emph{Close analogy} \\


How does the digestive system work? & How does weathering cause rocks to break? & \emph{Far analogy} \\

How does a solar panel work? & What happens during photosynthesis? & \emph{Far analogy} \\

What happens during photosynthesis? & How does a virus infect an animal? & \emph{Not analogy} \\

\hline
\end{tabular}
\caption{Examples of different types of analogies, from the top of our method (FMQ) ranking on ProPara.}
\label{tab:analogy_types_examples}
\end{table*}

One of the reasons for developing a metric for analogous similarity between paragraphs is to be able to \emph{retrieve} analogies from a large corpus ({\bf RQ1}). 

To find analogies, we wish to \emph{rank} all $\sim$76K possible pairs (over the 390 ProPara paragraphs), so that analogies rise to the top. 
%
We expect very few pairs of paragraphs to be truly analogous, while the number of pairs that might happen to have one or two strong entity matches could be significantly higher. Thus, we prefer a mapping involving more entities, even though their scores are not very strong. We chose a simple ranking formulation that balances between the number of mappings and their strength -- multiplying the number of mappings by the median similarity, $|\M|\cdot\textit{median}(\M)$.


To the best of our knowledge, there is no baseline that solves our task. 
We first compare our method,  {\bf FMQ}, 
to {\bf SBERT} \citep{reimers-2019-sentence-bert}, a well-known method to derive semantically meaningful sentence embeddings for tasks like semantic textual similarity. The ranking is based on cosine similarity between paragraph embeddings. 

The second baseline we use is a simpler variant of our method we call {\bf Find Mappings by Verbs (FMV)}. FMV is identical to FMQ, but when finding a mapping (Section \ref{subsec:findMappings}) we compute the similarity between the \emph{verbs of the questions} instead of between the questions themselves. As verbs represent  \emph{relations}, which are a core part of analogies, this baseline is meant to test the additional benefit of using the questions extracted by QA-SRL.

We rank the 76K possible pairs via all three methods. We annotate the top 100 pairs, as well as 40 pairs from all quartiles  (bottom, middle, 25\% and 75\%), resulting in a total of 260 annotated pairs from each list (702 unique). The main intersections are between FMQ and FMV (25\% top, 95\% bottom), and between FMQ and SBERT (11\% top).

\xhdr{Labels}
If the texts are not analogous to each other, we use the \textbf{Not analogy} label. 
Analogies are divided into \textbf{Self analogy} (entities and their roles are identical), \textbf{Close analogy} (a close topic, entities from a similar domain),  \textbf{Far analogy} (unrelated topics with different entities), and \textbf{Sub-Analogy} (only a part of one process is analogous to a part of the other; 
should contain at least two similar relations.) 
See Table~\ref{tab:analogy_types_examples} for examples of different analogy types. Notice that we only show the prompt, but annotators could see the full paragraphs.

\xhdr{Annotation}
We had an expert (member of our team) annotate the 702 \emph{unique} pairs from the three lists 
in a double-blind fashion. As this is not an easy annotation, we performed two checks to assess the clarity and consistency of our annotation scheme. 

First, another {expert} from our team (highly familiar with analogies) annotated a sample of the data (containing all labels), achieving 90\% agreement, {with Cohen’s Kappa of 0.74 for the 2-labels and 0.88 for the 5-labels.}
Next, we recruited 15 volunteer annotators (graduate students in CS, most with a basic knowledge of analogies). We first trained the annotators, showing them two examples for each label, along with the correct label and an explanation. Annotators discussed the examples with the experimenter.
We sampled from our expert's annotation 5 pairs for each label, resulting in 25 pairs of paragraphs. Each annotator received 5 pairs, s.t. each pair is assigned to 3 annotators. 


\begin{figure}[t]
\begin{centering}
\includegraphics[scale=.22]{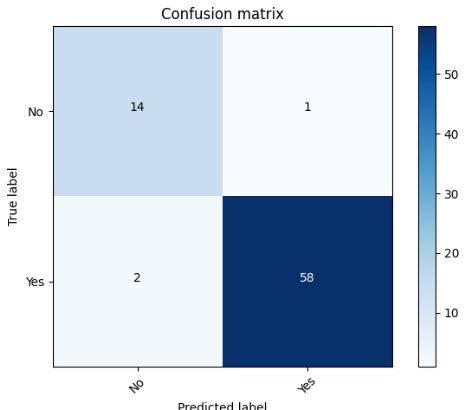}
\includegraphics[scale=.22]{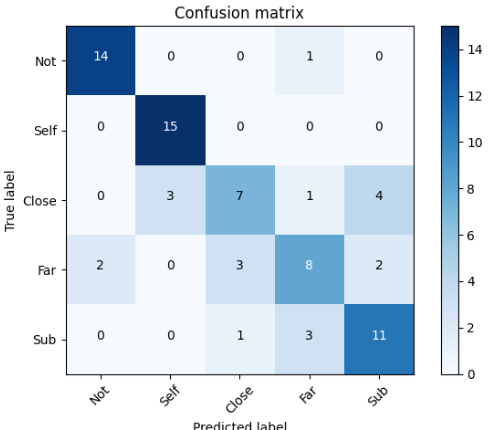}
\par\end{centering}
\caption{
Confusion matrices comparing 15 annotators to an expert annotator. \textbf{Left:} 2-labels granularity (analogy/not-analogy). \textbf{Right:} 5-labels (No, Self, Close, Far, Sub). We see very high agreement on binary labels, and that \emph{Self analogy} is the easiest for our annotators.}
\label{fig:conf_mat_annotators_consistency}
\end{figure}

When treating our expert's annotations as ground-truth, annotators' accuracy was 0.96 for the binary (analogy/non-analogy) task, and 0.73 for the 5-class task (27\% scored perfectly, 33\% had one mistake).
Figure~\ref{fig:conf_mat_annotators_consistency} shows the confusion matrices.
We can see a very high agreement in the binary task, which is most important in our setting, as our goal is to mine analogies; we have created the fine-grained labels for conducting more in-depth analysis of the found analogies.  
Out of the analogy types, we see that self-analogy was the easiest for annotators. Annotators also mentioned it was sometimes hard to distinguish between \emph{Far} and \emph{Sub} analogies.
{Fleiss Kappa is 0.82 for the 2-labels and 0.58 for the 5-labels.}
We conclude from the two sanity checks that our annotation schema is overall effective, and  the expert annotations can be used for evaluation.


\begin{table}
\centering
\begin{tabular}{llllll}
\hline
\textbf{Method} & \textbf{Not} &  \textbf{Sub} & \textbf{Self} & \textbf{Close} & \textbf{Far}\\
\hline
SBERT & 0 & 0 & 89 & 11 & 0 \\
FMV & 28 & 15 & 26 & 20 & 11\\
FMQ & 21 & 16 & 29 & 18 & 16\\
\hline
\end{tabular}
\caption{Different type of analogies found in top-100 of the ranking. SBERT found mostly {self-analogies}, while FMQ and FMV found more interesting analogies.}
\label{tab:methods_performance_analogies_types}
\end{table}

\begin{table}[t]
\centering
\begin{tabular}{llll}
\hline
\textbf{Method} & \textbf{P} &  \textbf{AP} & \textbf{NDCG} \\
\hline
FMV (@25) & 
$0.68$ & 
$0.36$ & 
$0.4$ \\
\hspace{0.8cm} (@50) & 
$0.72$ & 
$0.37$ & 
$0.41$ \\
\hspace{0.8cm} (@75) & 
$0.71$ & 
$0.36$ & 
$0.43$ \\
\hspace{0.8cm} (@100) & 
$0.72$ & 
$0.36$ & 
$0.43$ \\
\hline
\hline
FMQ (@25) & 
$\mathbf{0.96}$ & 
$\mathbf{0.5}$ & 
$\mathbf{0.57}$ \\
\hspace{0.8cm} (@50) & 
$\mathbf{0.84}$ & 
$\mathbf{0.43}$ & 
$\mathbf{0.52}$ \\
\hspace{0.8cm} (@75) & 
$\mathbf{0.77}$ & 
$\mathbf{0.39}$ & 
$\mathbf{0.47}$ \\
\hspace{0.8cm} (@100) & 
$\mathbf{0.79}$ & 
$\mathbf{0.4}$ & 
$\mathbf{0.49}$ \\
\hline
\end{tabular}
\caption{\emph{Precision@k}, \emph{AP@k} and \emph{NDCG@k}. FMQ is consistently better than FMV in all the three metrics.}
\label{tab:fmv_fmq_metrics_comparison}
\end{table}

 
\xhdr{Results}
%
All methods had zero analogies in the 25\%, middle, 75\% and bottom samples; the only analogies found were at the top-100. 
At the top, SBERT found 100\% analogies, FMQ  reached 79\% and FMV -- 72\%. At first glance, it seems like SBERT is winning. However, a closer look at analogy types (Table~\ref{tab:methods_performance_analogies_types}) shows it mostly finds \emph{self-analogies} --  paragraphs describing the same process (which makes sense, as it is designed for textual similarity), which are technically analogies, but not very interesting ones. On the other hand, FMQ and FMV did manage to find many far, close and sub-analogies, which is our goal. 

To estimate the prevalence of analogous pairs in the data, we randomly sampled 100 more pairs and had the expert annotate them, finding 3\% analogies (one sub-analogy, one self analogy and one far analogy), confirming the hardness of the task.

Table~\ref{tab:fmv_fmq_metrics_comparison} compares FMV and FMQ in terms of information retrieval metrics --  \emph{Precision (P)@k}, \emph{Average Precision (AP)@k} and \emph{Normalized Discounted Cumulative Gain (NDCG)@k}. For NDCG,  we defined gains of 0, 1, 2, 3, 4 for not, sub, self, close and far respectively. 
We can see  FMQ is consistently better than FMV in all  metrics, supporting our intuition that questions are more useful than verbs alone. See Appendix~\ref{subsec:fmq_fmv_top_rank_comparison} for full plots.

\subsection{Evaluating the Mappings}

In the previous section, we saw our method is able to identify analogies. However, we did not check that it indeed finds strong (true) mappings. In this section, we take a closer look at the mappings themselves and tackle {\bf RQ2} -- does our algorithm produce the correct mapping?


We chose 15 analogous pairs of paragraphs from ProPara, identified in the previous experiment (Section \ref{subsec:analogiesMining}), equally divided between close, self and far (we did not take sub-analogies as they are harder for the annotators, sometimes with more than one correct answer). We assigned one paragraph to each of our 15 volunteer annotators and asked them to find the correct mapping between the entities. 

{We note that choosing pairs of paragraphs from the previous experiment might introduce some bias. To evaluate mappings, we need pairs of analogous texts.  Randomly sampling and annotating ProPara pairs would be prohibitively expensive due to sparsity of analogies in the data; we do believe that analyzing a sample of the analogies found through our previous experiment, despite the potential bias, is still interesting and worth exploring.}

In addition, we  decided to try our algorithm on a different kind of data -- {\bf analogous stories} from cognitive-psychology literature {(which does not suffer from the potential bias mentioned above)}. 
We used the Rattermann and Keane problems \citep{gentner1993roles,ichienVerbalAnalogyProblem2020}, resulting in 19 pairs of stories. We filtered out stories with dialogue or at most two mappings, leaving us with 14 pairs. See Appendix~\ref{subsec:analogous_stories} for an example pair, along with our algorithm's mapping. We assigned these stories to 14 annotators (one story per annotator), and asked them to do the same as above.


We instructed the annotators to find mappings between entities, and emphasized that the mappings should be consistent and based on the roles entities play in the texts. We showed them two examples of correct mappings with explanations. One user provided an invalid mapping for ProPara, and we discarded his annotation.

We consider the annotators' labels as {ground truth}, and the algorithm's mappings  as {predictions}. We compare the performance of FMQ and FMV (Section \ref{subsec:analogiesMining}).
Again, to the best of our knowledge, there is no baseline for computing mappings. 
 


\xhdr{Results}
Table~\ref{tab:propara_and_stories_precision_recall_f1_score} shows {precision}, {recall} and {F1 score} for top-1 and top-3 solutions (from beam search).
%
%
%
Our method (FMQ) achieves higher {precision} and {recall} than FMV on both datasets. It has a very high agreement with the annotators (P@3 of 0.87 in ProPara and 0.94 in stories). We believe this is the result of the richer information provided by the questions (e.g., the answers to ``what is produced?'' and ``where is produced?'' are probably not analogous, despite the same verb).


We note that the results on the stories are better for both methods, which is probably due to the fact that the stories are written as analogies in the first place, with explicit parallels between them.

\xhdr{Recall analysis}
We analyzed FMQ's recall errors on stories and found that recurring sources of error include filtering ``Where'' questions, ``How'' questions and ``Be'' verbs. Refer to Appendix \ref{sec:ignore_qa} for a short discussion of our filtering design choices. These error patterns apply to ProPara as well. 

\begin{table}[t]
\centering
\begin{tabular}{lllll}
\hline
\textbf{Dataset} & \textbf{Method} & \textbf{P} &  \textbf{R} & \textbf{F1}\\
\hline
ProPara & FMV (@1) &  $0.48$ &  $0.33$ &  $0.39$  \\
 & FMQ (@1) & $\mathbf{0.82}$ & $\mathbf{0.64}$ & $\mathbf{0.72}$ 
 \\
 \hline
 & FMV (@3) & $0.58$ & $0.40$ & $0.47$ \\
 & FMQ (@3) & $\mathbf{0.87}$ & $\mathbf{0.67}$ & $\mathbf{0.76}$ \\
\hline
\hline
Stories & FMV (@1) & $0.64$ & $0.46$ & $0.54$ \\
 & FMQ (@1) & $\mathbf{0.88}$ & $\mathbf{0.68}$ & $\mathbf{0.77}$ 
 \\
 \hline
 & FMV (@3) &$0.73$ & $0.52$ & $0.61$ \\
 & FMQ (@3) & $\mathbf{0.94}$ & $\mathbf{0.76}$ & $\mathbf{0.84}$ \\
\hline
\end{tabular}
\caption{Evaluating the mappings of our method (FMQ) and FMV in terms of  {Precision (P)}, {Recall (R)} and {F1 score}. We compare the metrics on top-1 and top-3 solutions  of beam search. FMQ outperforms FMV across all metrics. }
\label{tab:propara_and_stories_precision_recall_f1_score}
\end{table}


\label{subsec:mappingsEvaluation}

\subsection{Robustness to Paraphrases}

Our method heavily relies on the way the input texts are phrased.
In this experiment, we examine robustness to \emph{paraphrasing} ({\bf RQ3}). 

\xhdr{Automatic paraphrases}
We focus on the ProPara dataset. We chose ten paragraphs which are not analogous to each other and generated four paraphrases using \emph{wordtune}, a large-scale language model\footnote{\url{https://wordtune.com}}: two expansions and two abridgments (examples in Appendix~\ref{subsec:wordtune}). This results in 50 paragraphs, or 1225 pairs. 
{We note that 
~80\% of paraphrased sentence pairs contained different verbs.}

We labeled the 100 pairs that came from the same original paragraph as an analogy (in fact, they are \emph{self-analogies}), and the rest as non-analogy. Then, we rank all pairs via SBERT, FMV and FMQ.  Table~\ref{tab:robustness_precision} shows \emph{precision@k} for top 100 (full plot in Appendix~\ref{subsec:robustness_precision}).
Both FMV and FMQ achieve very high precision (near-perfect for the first 50, and very high at 100), demonstrating that our method is relatively \emph{robust} to \emph{paraphrases}. 
As a reality check, SBERT achieves near-perfect results, which makes sense as these are self-analogies, and it is designed for textual similarity. 

\xhdr{Responses to the same prompt}
In the ProPara dataset, the same prompt is sometimes given to multiple authors. We now explore whether our algorithm can recognize those (self) analogies. Again, we take ten non-analogous paragraphs given to at least five authors, and randomly choose five authors for each, resulting in 1225 pairs of paragraphs, with 100 labeled as analogies.
Unlike the previous experiment, the labels here are much \emph{noisier}, as authors given the same prompt can focus on different aspects or granularity, resulting in non-analogous paragraphs (e.g., when describing the human life cycle, some authors mention zygotes and embryos, and others mention teenagers and adults. See Appendix~\ref{subsec:same_prompt}).

Table~\ref{tab:robustness_precision} shows \emph{precision@k} results (full plot in Appendix~\ref{subsec:robustness_precision}).
Precision is reasonable, but lower than automatic paraphrases. The labels themselves are noisy, so it is harder to assess performance. 
{Note that \emph{recall@100} = \emph{precision@100} in both automatic paraphrases and responses to the same prompt, as there are 100 true positives.}

\xhdr{Error analysis}
Looking at our model's false negatives, we see mostly pairs of paragraphs describing the same topic from different points of view, which is really a mistake in the ground truth (see example in Appendix~\ref{subsec:no_similar_verbs_example}). We note that some annotators reported it was impossible to notice two paragraphs were self-analogies without seeing the (identical) prompts.
SBERT is not much affected by this, as it looks at the entire text, while our methods are ``blind'' to the entities.
Another interesting source of false negatives is mistakes introduced by wordtune (e.g., expanding 
``the water builds up'' to
``Nitrates build up in the body of the water'').
For false positives, we identify several sources of error, such as non-analogous texts with similar verbs, QA-SRL handling of phrasal verbs (``take care'', ``take off''), repeating verbs, and extraction issues (for example, the sentence ``Water, ice, and wind hit rocks'' lead to singleton entities and ``water, ice, and wind'', resulting in double-counting).


\begin{table}
\centering
\begin{tabular}{lllll}
\hline
\textbf{Method} & \textbf{P@25} &  \textbf{P@50} & \textbf{P@75} & \textbf{P@100} \\
\hline
FMV & 
$1.00$ & 
$0.98$ & 
$0.93$ & 
$0.83$ \\
FMQ & 
$1.00$ & 
$0.98$ & 
$0.89$ & 
$0.74$  \\
SBERT & 
$1.00$ & 
$1.00$ & 
$1.00$ & 
$1.00$  \\
\hline
\hline
FMV & 
$0.64$ & 
$0.48$ & 
$0.33$ & 
$0.28$ \\
FMQ & 
$0.64$ & 
$0.5$ & 
$0.44$ & 
$0.38$  \\
SBERT & 
$1.00$ & 
$1.00$ & 
$0.97$ & 
$0.92$  \\
\hline
\end{tabular}
\caption{\emph{Precision@k} (P@k) in the \emph{automatic paraphrasing} (top) and the \emph{same prompt} task (bottom). Our methods achieve very high results in the \emph{paraphrasing} task. The results in the \emph{same prompt} task are moderately high, perhaps due to noisy ground truth (pairs about the same topic erroneously tagged as analogies). SBERT performs almost perfectly, as it sees the entire text, while our methods are blind to the entities.}
\label{tab:robustness_precision}
\end{table}



\label{subsec:robustnessEvaluation}

\section{Related Work}

\xhdr{Computational analogy} 
Classical analogy-making approaches are typically categorized as symbolic, connectionist and hybrid. See \citep{french2002computational,mitchell2021abstraction,gentner2011computational} for a comprehensive review.


In the NLP community, most work focused on simple word analogies. This area has gained popularity after showing that word embeddings can model some relational similarities in terms of word vector offsets (\emph{``king - man + woman = queen''} \citep{mikolov2013distributed}). Recent works \citep{linzen-2016-issues, ushio2021bert, NEURIPS2020_1457c0d6} show that the offset method only works for some types of relations, struggling with complex, abstract ones.

In a different setting, LRME \citep{turney2008latent} took as input two sets of entities (base, target) and tried to extract a common relational structure between them, mining relations from a large web corpus. Unlike our work, their setting focused on commonsense relations (e.g., electrons revolve around the nucleus), and could not handle either procedural texts (where entities go through multiple stages) or relations that are situation-specific. Also, LRME requires entities and relations to be expressed exactly the same across domains, making it brittle.

Other works combined NLP and crowds to find analogies between products \citep{kittur2019scaling, hope2017accelerating}. The models do not explicitly define entities and relations, but instead extract \emph{schemas} and match based on them.

\xhdr{Aligning texts}
Multi-text applications often need to model redundancies across texts. 
There has been much work in this area, exploiting graph-based representations \citep{cui2020inducing, yu2021graph}, QA-SRL \citep{brook-weiss-etal-2021-qa}, and other semantic structures. This work is in similar spirit to ours, but the crucial difference is that in our task the algorithm is blind to entities, and only focuses on aligning the relations.



\xhdr{Procedural texts}
%
%
Understanding procedural text has a long history in NLP, with a lot of work focusing on event extraction, tracking what happens to entities throughout the text \citep{Henaff2017TrackingTW,Tandon2018ReasoningAA,Bosselut2018SimulatingAD}. 
More recently, \citep{Dalvi2019EverythingHF} extended these frameworks, to also understand \emph{why} some actions need to happen before others.
To the best of our knowledge, the task of finding analogous mapping between procedural texts is novel.



\remove{
\xhdr{Cognitive psychology stories}
The Cognitive Psychology Stories Problem Sets is described in \citet{ichienVerbalAnalogyProblem2020}. We used the Rattermann set \citep{gentner1993roles}. This inventory was created to examine the role of analogical similarity in accessing episodes in human memory. 
In our work with use the dataset in order to evaluate our algorithm mappings performance on analogous stories, which is a totally different kind of texts than ProPara.
}

\section{Conclusions and Future Work}

Analogies can facilitate learning and problem-solving, helping people apply their prior experience to new situations. 
Much research has suggested that analogy-making is necessary for AI systems to robustly generalize and adapt to new contexts.

%
%
In this work we explored a complex, challenging analogy-finding setting: our input is a pair of natural language procedural texts, describing a situation or a process. 
We presented a novel, scalable and interpretable method to extract entities and relations from the text and find a mapping between entities across the domains.
We show that our method successfully identifies the correct mappings between the domains for both procedural texts from ProPara, and stories from cognitive-psychology literature. We demonstrate our method can be used to mine analogies from ProPara, including far, non-trivial analogies. Lastly, we show that our method is robust to \emph{paraphrases} the input texts.

In the future, we plan to improve our relation extraction and augment the text with commonsense knowledge to account for relations that do not appear explicitly in the text. We also plan to extend our algorithm to take the \emph{order} of actions into account, and to apply our method to new domains, such as \emph{legal texts} and \emph{recipes}. Two particularly interesting applications are (1) \emph{education}, where analogies can help a teacher explain a complex concept, and (2) \emph{computer-assisted creativity}, where engineers and designers could find inspiration in distant domains.


We release our dataset, including the found analogies and their mappings {(\url{ https://github.com/orensul/analogies_mining})}. We hope this work will encourage the development of novel NLP methods for computational analogy.


\paragraph{Acknowledgements}

{We would like to thank Aniket Kittur and Joel Chan for their valuable feedback and ideas, the Hyadata Lab and NLP-HUJI members for their thoughtful remarks, and all the participants in our experiments for their efforts. We also thank the anonymous reviewers for their constructive comments. This work was supported by the European Research Council (ERC) under the European Union’s Horizon 2020 research and innovation programme (grant no. 852686, SIAM).}

\section*{Limitations}

\begin{compactitem}
    \item \textbf{Relation extraction}: as discussed in Section \ref{subsec:findMappings}, QA-SRL misses some relations (e.g., those that are expressed across multiple lines). This reduces the effectiveness of our method. For this reason, we also expect our method to work best on more technical descriptions (where there are actions and entities that can be tracked), and less on paragraphs with a narrative style.
    \item \textbf{Insensitivity to order of actions}: our method does not take the order in which actions took place into consideration. For example, a sequence of actions and its reverse sequence would look analogical to the model. 
    \item \textbf{Handling of phrasal verbs}: QA-SRL does not handle phrasal verbs well, reducing phrases such as `take care'' and  ``take off'' to the verb ``take'' (``what takes something?''). 
    \item \textbf{Language}: Our datasets contain solely English texts. The results may differ in other languages.
\end{compactitem}

\section*{Ethics Statement}

To the best of our knowledge, the datasets used in our work do not contain sensitive information. The cognitive psychology dataset is somewhat antiquated, and might be considered to reinforce some stereotypes; we note we do not train any model on this data, but only use it for evaluation.


\newpage




\bibliography{anthology,custom}
\bibliographystyle{acl_natbib}

\appendix
\label{appendix}

\section{QA-SRL Ignore QA criteria}
Here we explain our criteria for ignoring QA which we parsed from the QA-SRL model.
\begin{itemize}

\item When the probability of the question (according to QA-SRL) is less or equal to our question probability threshold (see ~\ref{subsec:other_params} for the chosen threshold). 

\item When the probability of the answer (according to QA-SRL) is less or equal to the answer probability threshold (see ~\ref{subsec:other_params} for the chosen threshold). It means that the probability this is the correct span is too low.

\item When the question is not one of ``what'', ``who'' and ``which''. The rationale is that we focus on questions most likely to capture useful relations for our task.

\item When the question's verb is ``be''. The rationale is that ``be'' is not indicative enough. For example, if we have ``X was something'' on one text and ``Y was something'' on the other text, it does not indicate that X and Y play similar roles.

\item When the answer contains a verb or does not contain a noun. 
In this case, it does not an entity according to our definition.

\item When the answer is a pronoun.

\end{itemize}
\label{sec:ignore_qa}

\section{Parameters Fine-tuning}

Here we discuss how we choose the value for different parameters which are used in the experiments. We note our goal in this paper was to come up with a proof-of-concept system, and further parameter tuning might improve the results. 

\subsection{Cosine Similarity Threshold} 
\label{subsec:cosine_threshold}
\begin{figure*}[t]
\begin{centering}
\includegraphics[scale=.4]{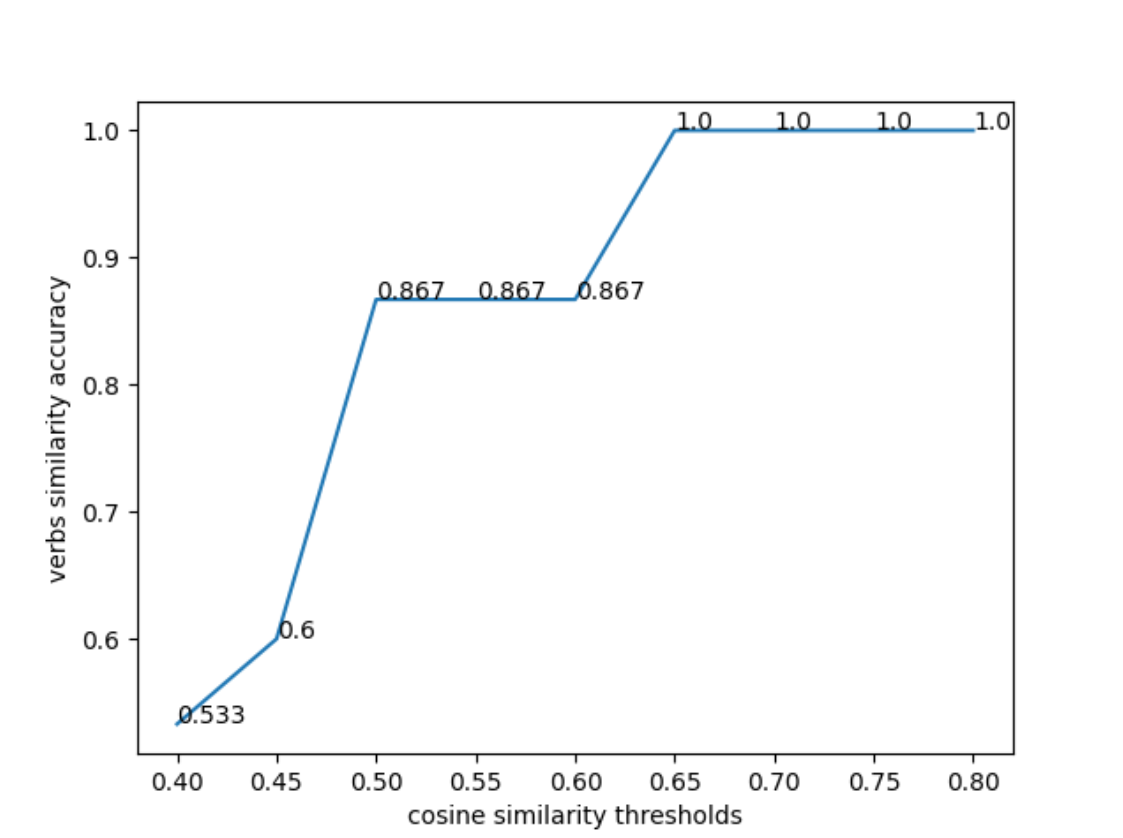}
\includegraphics[scale=.4]{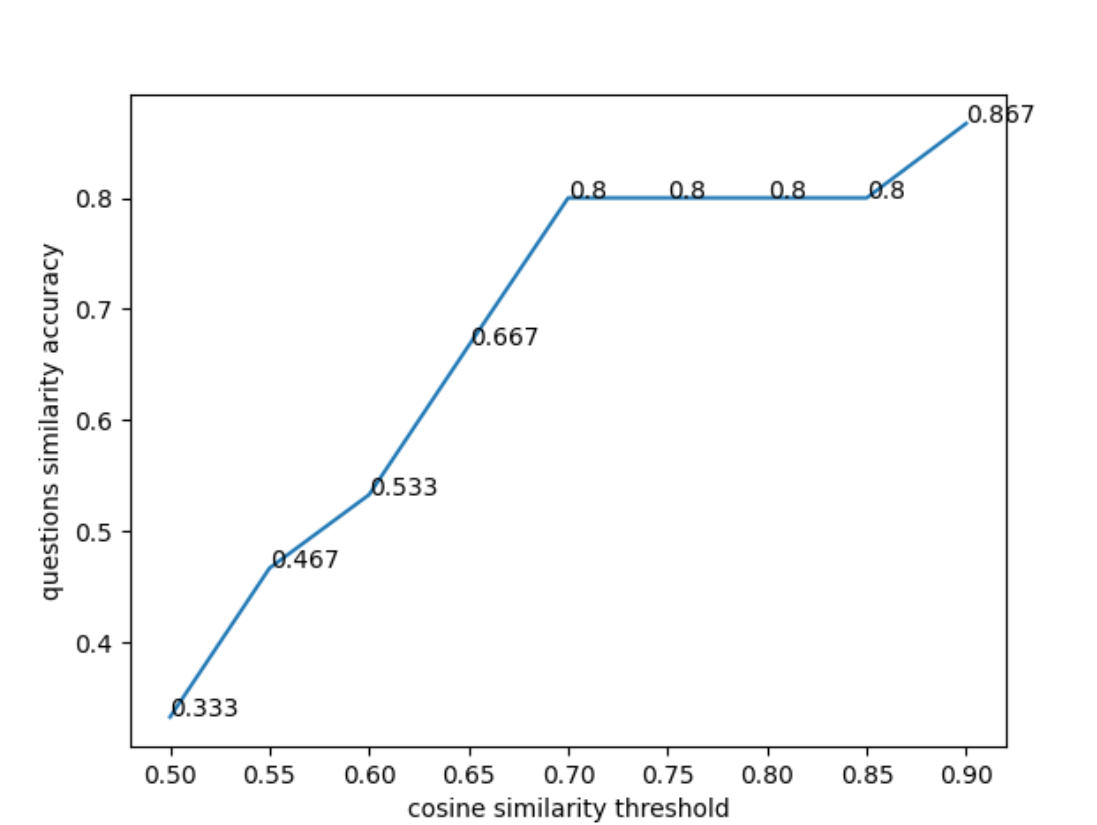}
\par\end{centering}
\caption{\emph{Verbs} similarity accuracy (on the left) and \emph{questions} similarity accuracy (on the right) for different cosine similarity thresholds. We can see that a threshold of 0.5 for the \emph{verbs} and 0.7 for the \emph{questions} achieves accuracy of at least 0.8.}
\label{fig:cosine_similarity_thresholds}
\end{figure*}

To determine the best cutoff for cosine similarity threshold between questions (FMQ) or between verbs (FMV), we did the following: We sampled 15 pairs of verbs (from ProPara paragraphs) in every range of threshold (intervals of 0.05 from 0 to 1 for \emph{questions} and for \emph{verbs}), then manually labeled the pairs of verbs as similar or not, and chose the threshold of \textbf{0.5}. The threshold was chosen to balance between precision (percentage of correct samples from all samples passing the threshold) and  an estimation of recall (percentage of correct samples from all correct samples that do manage to pass the threshold), computed using samples from all intervals.  For the similarity of the questions we did the same process, but instead of verbs, we sampled questions. We found that \textbf{0.7} is the best cutoff. 
See Figure~\ref{fig:cosine_similarity_thresholds} for different thresholds and their similarity accuracy.

\subsection{Other Parameters}
\label{subsec:other_params}
We choose the values for the following parameters by \emph{manual fine-tuning}.
For the QA-SRL settings, we set the \emph{answer probability threshold} to \textbf{0.05} (range checked: 0.0-0.15) and \emph{question probability threshold} to \textbf{0.1} (range checked: 0.0-0.15, intervals of 0.05), optimizing for F1-score. We set the Agglomerative clustering \emph{distance} parameter to be \textbf{1.0} (range checked: 0-2 with a step size of 0.5), optimizing for cluster purity compared to a ground-truth clustering on several examples. 
For the SBERT embedder model, we used the pre-trained model \emph{msmacro-distilbert-base-v4} \citep{reimers-2019-sentence-bert}. When a complete \emph{relation} was found, we add $\alpha$ of \textbf{1.0} (range checked: 0-2 with a step size of 1) 
for both pairs of entities in base and target, again optimizing on several left-out examples. We set the beam search size to \textbf{7} (range checked: 1-10). 

\label{sec:params_fine_tuning}

\section{Analogies Mining}

\subsection{FMQ / FMV top@k comparison}
\label{subsec:fmq_fmv_top_rank_comparison}
\begin{figure*}[t]
\begin{centering}
\includegraphics[scale=0.35]{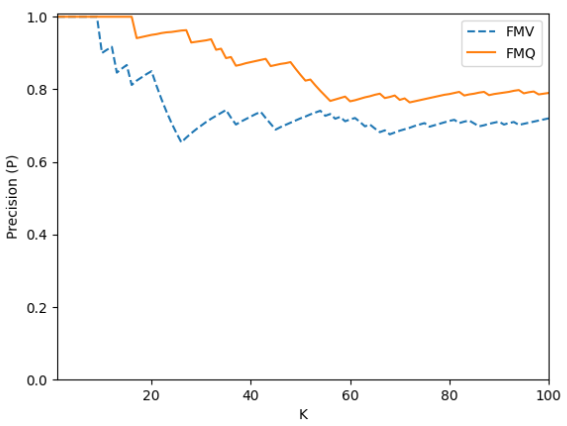}
\includegraphics[scale=0.35]{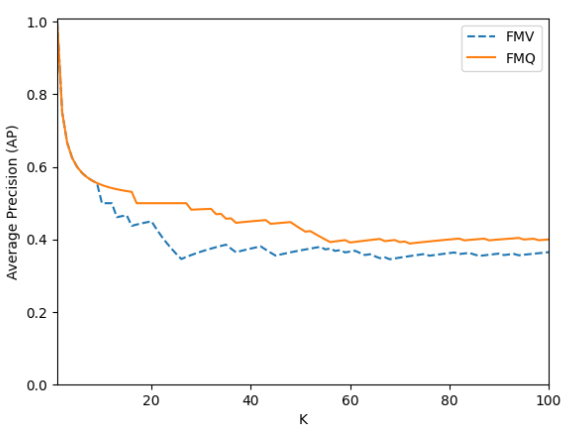}
\includegraphics[scale=0.35]{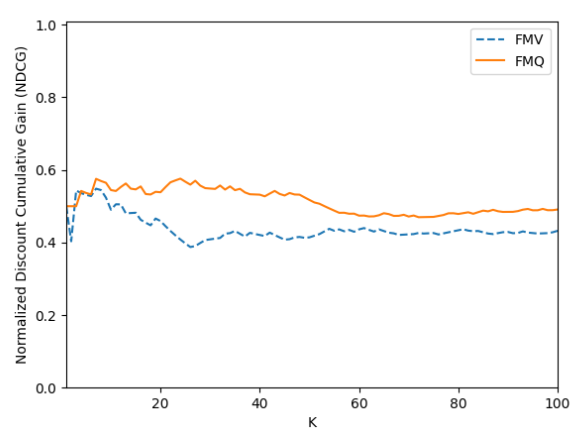}
\par\end{centering}
\caption{Comparison between FMQ and FMV methods in terms of \emph{Precision (P) @k}, \emph{Average Precision (AP) @k} and \emph{Normalized Discounted Cumulative Gain (NDCG) @k} ($k$ from 1 to 100). As we can see, FMQ is consistently better than FMV in all the three metrics.}
\label{fig:fmv_fmq_comparison}
\end{figure*}

Here we compare our method (FMQ) and FMV on the top of their lists according to information retrieval metrics.
See Figure~\ref{fig:fmv_fmq_comparison} for three plots which compare between FMV and FMQ methods on the different metrics.

\label{sec:analogies_mining}

\section{Mappings Evaluation Examples}

\subsection{Pair of ProPara paragraphs} 
\label{subsec:propara_paragraphs}
\begin{figure*}[t]
\includegraphics[scale=.32]{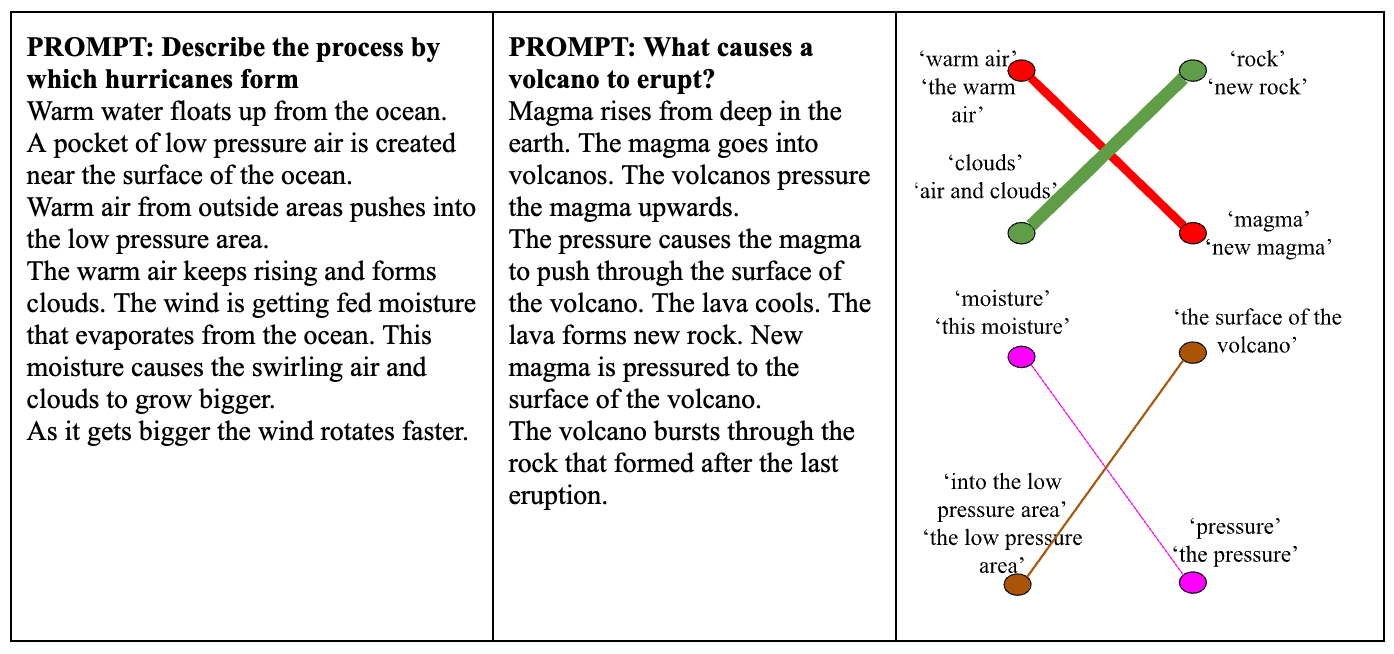}
\caption{Left and Middle: two \emph{procedural texts} paragraphs from ProPara, describing hurricanes (base) and volcano eruptions (target). 
Right: Our algorithm’s output. The nodes are entities (clusters of text spans; for the sake of presentation, we show up to two spans). Edge width represents similarity between entities in terms of the roles they play in the text. According to the annotation, all mappings are correct.}
\label{fig:pair_of_paragraphs_solution}
\end{figure*}

See Figure~\ref{fig:pair_of_paragraphs_solution} for two example paragraphs from ProPara and our algorithm's mapping.

\subsection{Pair of analogous stories} 
\label{subsec:analogous_stories}

\begin{figure*}[t]
\includegraphics[scale=.32]{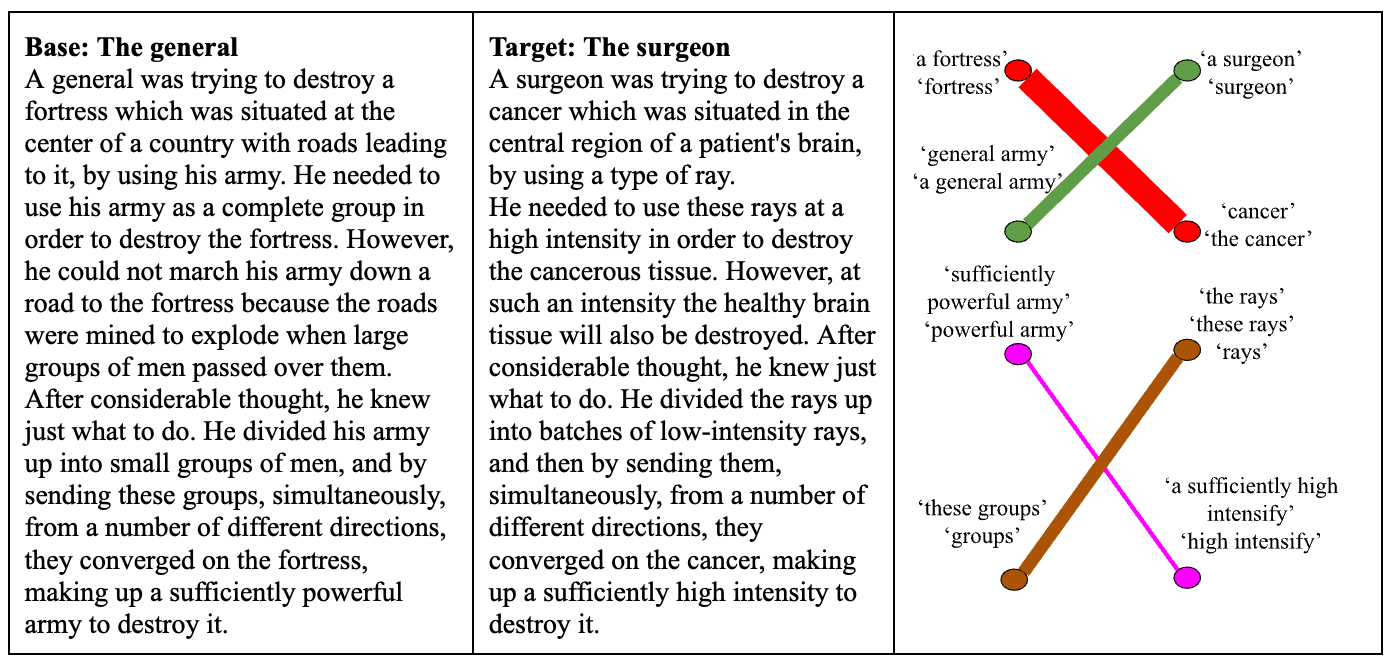}
\caption{Left and Middle: two analogous \emph{stories} from cognitive psychology literature, describing a general trying to attack a fortress (base) and the surgeon trying to destroy a tumor (target).
Right: Our algorithm’s output. The nodes are entities (clusters of text spans; for the sake of presentation, we show up to two spans). Edge width represents similarity between entities in terms of the roles they play in the text. 
According to the annotation, the \emph{precision} is perfect, but the \emph{recall} is not as the following mappings are missing: 
``center of a country'' to  ``central region of a patient's brain'' and ``roads'' to ``healthy tissue'' (which we miss is because we filter ``where'' questions).}
\label{fig:pair_of_stories_solution}
\end{figure*}

See Figure~\ref{fig:pair_of_stories_solution} for the texts and our algorithm output on the general/surgeon analogous stories.

\label{sec:mappings_eval_examples}

\section{Algorithm Robustness}

Here we show examples of automatic paraphrases made by wordtune as well response of different authors to the same prompt. We show the results of our algorithm's robustness. Finally, we analyze the main errors sources. 

\subsection{Automatic Pharaphases}
\label{subsec:wordtune}
We use the tool \emph{wordtune}, which was developed by AI21 for our \emph{paraphrases'} evaluation on paragraphs from ProPara, as part of our robustness evaluation (\ref{subsec:robustnessEvaluation}). \emph{Wordtune} has several features, some of them are \emph{wordtune short} for rewriting the text in a shorter way, and \emph{wordtune expand} for rewriting the text with longer sentences. See Figure~\ref{fig:wordtune} for an example.

\begin{figure*}[t]
\fbox{\parbox[c]{15.5cm}{
\textbf{Original paragraph:} How do lungs work? \\
You breathe air in. Air enters bronchial tubes. 
Air is then split into the bronchioles of each lung. 
Bronchioles have alveoli which are tiny air sacs.
Alveoli is surrounded by many small blood vessels. 
Oxygen passes through alveoli into blood vessels. 
Blood leaves the lungs as it travels through the body. 
Blood carries carbon dioxide back to the lungs. 
Carbon dioxide released when you exhale.
}}
\fbox{\parbox[c]{15.5cm}{
\textbf{Wordtune expand:} \\
When you breathe in, you are taking in air. 
Through your bronchial tubes, air enters your lungs. 
After the air has passed through the bronchial tubes, 
it is divided into the bronchioles of each lung. 
Alveoli, which are tiny sacs of air, are situated 
in the bronchioles. The alveoli are surrounded by 
a big number of small blood vessels. It is through 
these blood vessels that oxygen moves into the alveoli. 
In the course of its journey through the body, 
the blood enters through the lungs. When blood 
returns to the lungs, it takes carbon dioxide along with it. 
It is this carbon dioxide that is released when you breathe out.
}}
\fbox{\parbox[c]{15.5cm}{
\textbf{Wordtune short:} \\
Breathing air in. Bronchial tubes obtain air. Lungs split air into bronchioles. Alveoli are tiny air sacs in the bronchioles. Small vessels nearby alveoli. Alveoli grab oxygen to blood vessels. As blood passes through the body, it leaves the lungs. CO2 is carried by blood to the lungs.
CO2 is discharged when you breathe out.
}}
\caption{An example of \emph{paraphrases} -- one long (top) and one short (bottom) made by \emph{wordtune} for a paragraph describing how lungs work. }
\label{fig:wordtune}
\end{figure*}

\subsection{Responses to the Same Prompt}
\label{subsec:same_prompt}
In addition to paraphrases, we also want to challenge our system with different text versions of different authors on the same topic. This is a part of the robustness evaluation (\ref{subsec:robustnessEvaluation}). To do so, we take four different versions for one original paragraph. See Figure~\ref{fig:same_prompt} for an example.

\begin{figure*}[t]
\fbox{\parbox[c]{15.5cm}{
\textbf{Original paragraph:} Describe the life cycle of a human \\
A human baby develops in the womb of the mother.
After 9 months in the womb the baby is born. 
It is an infant.
The infant is dependent on its parents for everything.
It drinks mother's milk for nourishment.
From 3-8 years old the child is in early childhood.
Adolescence is from roughly 9-18 years old.
During adolescence the child is growing rapidly and maturing sexually.
At 18 years, the child becomes an adult.
Adults can reproduce and have babies.
}}
\fbox{\parbox[c]{15.5cm}{
\textbf{V1:} \\
A human is born.
The human is an infant.
The infant grows into a toddler.
The toddler grows into a child.
The child grows into a teenager.
The teenager grows into an adult.
The adult grows old.
The human dies.

}}
\fbox{\parbox[c]{15.5cm}{
\textbf{V2:} \\
A human is born.
The human is a child and learns.
The human child grows into an adult.
The adult uses its skills to survive.
The human starts a new family and propogates.
The human grows old.
The human dies.
}}
\fbox{\parbox[c]{15.5cm}{
\textbf{V3:} \\
A zygote is formed via sexual reproduction.
This zygtoe grows in the womb to become a fetus.
After a typical 9-month period, a human is born.
The human is an infant at this stage.
The infant becomes a toddler, and learns to walk and speak.
The toddler becomes a child.
The child becomes a teenager after undergoing puberty.
The teenager grows into an adult.
The adult hits a peak, and development stops.
Old age and eventually death occur.

}}
\fbox{\parbox[c]{15.5cm}{
\textbf{V4:} \\
A sperm fertilizes an egg.
The egg forms into a fetus.
9 months passes as the fetus grows into an infant.
The infant is born.
The baby begins to grow into an adolescent.
The adolescent turns into a young adult.
The young adult learns and grows into a fully mature adult.
}}
\caption{An example of different ProPara paragraphs with the same \emph{prompt} (written by different authors) describing the human life cycle. We can see authors focus on different aspects or granularity (e.g., some authors mention zygotes and embryos, and others mention teenagers and adults).}
\label{fig:same_prompt}
\end{figure*}

\begin{figure*}[t]
\begin{centering}
\includegraphics[scale=.35]{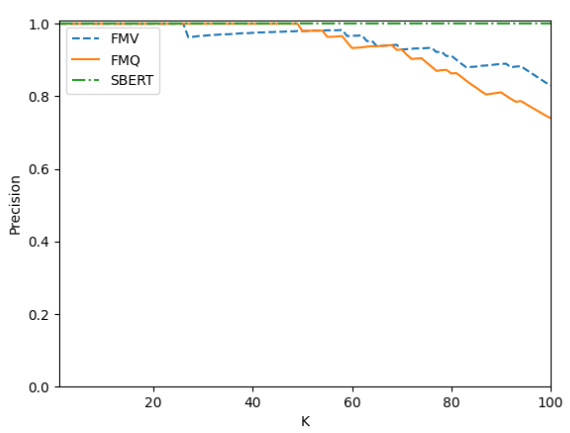}
\includegraphics[scale=.35]{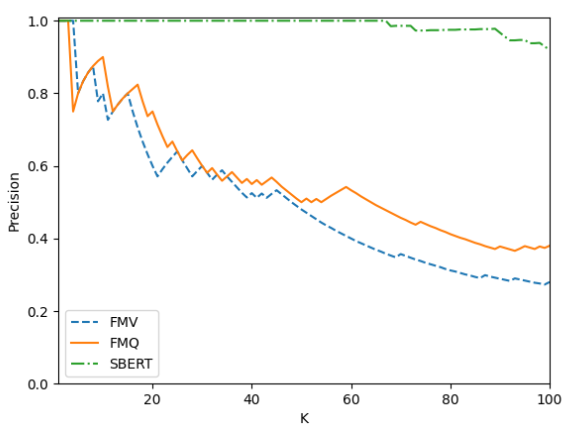}
\par\end{centering}
\caption{Left: \emph{Precision@k} in the \emph{automatic paraphrasing} task. Right: \emph{Precision@k} in the \emph{same prompt} task. Our methods achieve very high results in the \emph{paraphrasing} task. The results in the \emph{same prompt} task are moderately high, perhaps due to the noisier setting (many pairs tagged as analogies in our ground truth are about the same topic but not actually analogous). SBERT was able to perform almost perfectly, as it looks at the entire text, while our methods are blind to the entities.}
\label{fig:robustness_precision}
\end{figure*}

\subsection{Robustness Precision}
\label{subsec:robustness_precision}
See Figure~\ref{fig:robustness_precision} for two plots of  \emph{precision@k} for $k$ between 1-100 of the top-ranked pairs of paragraphs for \emph{paraphrases} and same \emph{prompt} evaluations.

\begin{figure*}[t]
\fbox{\parbox[c]{15cm}{
\textbf{V1: how does internal combustion engine work?} \\
Air and fuel are \textbf{used} in the internal combustion engine. 
In an enclosed chamber, a mixture of air and fuel is \textbf{injected}. 
The mixture \textbf{ignites} and \textbf{turns} a piston that \textbf{pumps} up and down. This piston is \textbf{connected} to a crankshaft which \textbf{rotates} to \textbf{provide} the power. 
The burned gas is \textbf{pushed} out of the chamber. }}

\fbox{\parbox[c]{15cm}{
\textbf{V2: how does internal combustion engine work?} \\
The piston \textbf{moves} down. Gasoline and air \textbf{go} 
into the engine. The piston \textbf{moves} back up. 
The gasoline and air are \textbf{compressed}. 
The spark plug \textbf{emits} a spark. 
The gasoline \textbf{explodes}. The explosion \textbf{forces} 
the piston down. The exhaust valve \textbf{opens}. 
Exhaust \textbf{goes} to the tailpipe. 
}}
\caption{An example of pair of paragraphs with the same prompt (written by different authors) which is ranked in the top 100 by SBERT (score of 0.54), but getting a zero score by our method, as these texts describe the same process (``How does internal combustion engine work'') from a different point of view and with no \emph{similar verbs} (verbs are in bold).}
\label{fig:no_similar_verbs_example}
\end{figure*}


\subsection{Same Prompt, No Similar Verbs}
\label{subsec:no_similar_verbs_example}
See Figure~\ref{fig:no_similar_verbs_example} for two paragraphs with the same prompt, but with different verbs.

\label{sec:algorithm_robustness}

\end{document}